\def\year{2022}\relax
\documentclass[letterpaper]{article} 
\usepackage{aaai22}  
\usepackage{times}  
\usepackage{helvet}  
\usepackage{courier}  
\usepackage[hyphens]{url}  
\usepackage{graphicx} 
\def\UrlFont{\rm}  
\usepackage{natbib}  
\usepackage{caption} 
\DeclareCaptionStyle{ruled}{labelfont=normalfont,labelsep=colon,strut=off} 
\usepackage{algorithm}
\usepackage{algpseudocode}

\usepackage{newfloat}
\usepackage{listings}
\floatstyle{ruled}
\newfloat{listing}{tb}{lst}{}
\floatname{listing}{Listing}
\title{Online Adaptation of Neural Network Models by Modified Extended Kalman Filter for Customizable and Transferable Driving Behavior Prediction}
\author{
    Letian Wang\textsuperscript{\rm 1}, Yeping Hu\textsuperscript{\rm 2}, Changliu Liu\textsuperscript{\rm 1}\\
}
\affiliations{
    \textsuperscript{\rm 1}Carnegie Mellon University, Pittsburgh, PA, USA 15213\\
    \textsuperscript{\rm 2}University of California, Berkeley, Berkeley, CA, USA 94720\\

    letianwang0@gmail.com, yeping\_hu@berkeley.edu, cliu6@andrew.cmu.edu
}

\usepackage{bibentry}

\begin{document}

\maketitle

\begin{abstract}

High fidelity behavior prediction of human drivers is crucial for efficient and safe deployment of autonomous vehicles, which is challenging due to the stochasticity, heterogeneity, and time-varying nature of human behaviors. On one hand, the trained prediction model can only capture the motion pattern in an average sense, while the nuances among individuals can hardly be reflected. On the other hand, the prediction model trained on the training set may not generalize to the testing set which may be in a different scenario or data distribution, resulting in low transferability and generalizability. In this paper, we applied a $\tau$-step modified Extended Kalman Filter parameter adaptation algorithm (MEKF$_\lambda$) to the driving behavior prediction task, which has not been studied before in literature. With the feedback of the observed trajectory, the algorithm is applied to neural-network-based models to improve the performance of driving behavior predictions across different human subjects and scenarios. A new set of metrics is proposed for systematic evaluation of online adaptation performance in reducing the prediction error for different individuals and scenarios. Empirical studies on the best layer in the model and steps of observation to adapt are also provided. 

\end{abstract}

\section{Introduction}

As more autonomous vehicles are being deployed on the road, safety is the most essential aspect to consider especially when interacting with other dynamic road entities. Being able to make high-fidelity future motion predictions for surrounding entities can greatly improve the safety level of autonomous vehicles, since actions can be generated accordingly to avoid potentially dangerous situations. 

The challenges for building accurate prediction models lie in following aspects. Firstly, the prediction models are usually trained with a limited pool of real human motion data. Though able to learn the average behavior pattern in the data, the model is incapable to capture the nuances among heterogeneous individuals. Secondly, due to the stochastic and time-varying nature of human behaviors, the same individual may also exhibit different behaviors in different circumstances, which requires the prediction models to comprise dynamic parameters rather than fixed ones. Thirdly, although a trained model typically performs well on the training set, performance can drop significantly in a slightly different test scenario or under a slightly different data distribution. 

To address these problems, this paper extends an existing hierarchical prediction model~\cite{wang2021hierarchical} for human driving behavior to include adaptability in real time. The pipeline of the framework is as follows: 1) learning the hierarchical prediction model via real human driving data, which consists of a semantic graph network for intention prediction and an encoder decoder network for trajectory prediction. 2) executing real-time $\tau$-step online adaptation for the trajectory prediction by the modified Extended Kalman Filter algorithm (MEKF$_\lambda$).

The contribution of the paper is as follows: 1) extending a hierarchical prediction model to be adaptable to account for heterogeneous and stochastic behaviors among different agents and scenarios. 2) applying MEKF$_\lambda$ online adaptation algorithm to a more complex problem and demonstrate its usefulness. 3) proposing a new set of metrics for systematic evaluations of online adaptation performance. Such metrics can be widely used to evaluate the performance of online adaptation in regardless of models and problems. Via these metrics, we also provided empirical studies on the best layer and the best steps of observation to adapt.

\section{Related Works}
Various methods have been proposed to make driving behavior predictions, including traditional method such as hidden markov models~\cite{deo2018would}, gaussian process~\cite{9357407}, and learning-based methods such as recurrent neural networks~\cite{park2018sequence}, and graph neural networks~\cite{hu2020scenario}. In terms of prediction objective, some works only focus on predicting intentions~\cite{hu2018framework} or trajectories~\cite{ma2019trafficpredict}, while some other works make both intention and trajectory predictions in a hierarchical manner~\cite{wang2021hierarchical}. While learning-based methods have shown greater expressiveness with little human effort, taking both intention and trajectory predictions into account could simplify learning and make the model more interpretable. A detailed survey can be found in~\cite{rudenko2020human}.

To address individual differences, many works have been utilizing feedback from historic observations to make adaptable predictions: adapting reward function parameters to make driving-style-aware prediction via bayesian inference algorithm~\cite{wang2021socially}, adapting the last layer of a feedforward neural network using the recursive least square parameter adaptation algorithm (RLS-PAA)~\cite{cheng2019human}, adapting the last layer of a recurrent neural network using RLS-PAA~\cite{8814238}, and adapting any layer of recurrent neural network using modified Extended Kalman Filter algorithm (MEKF$_\lambda$)~\cite{abuduweili2021robust}.

To account for heterogeneous behaviors among agents and scenarios, this paper extends the hierarchical prediction model~\cite{wang2021hierarchical} to include adaptability. The model conducts both intention and trajectory prediction on real human driving data, which is a more complicated model addressing more a complex problem compared to previous works on online adaptation. Also, a new set of metrics is proposed for systematic evaluation of online adaptation.

\begin{figure}[t!]
    \centering
    \includegraphics[width=0.5\textwidth]{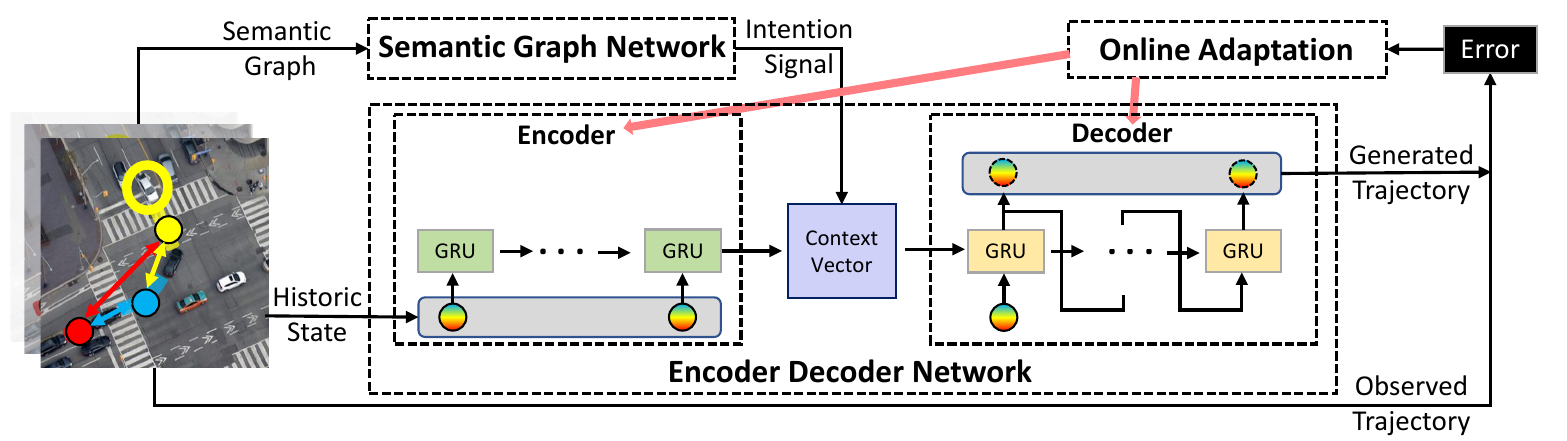}
    \caption{Network architecture in our framework.}
    \label{fig:architecture}
\end{figure}

\section{Problem Formulation}
Our goal is to generate more accurate driving behavior predictions in multi-agent traffic-dense scenarios. Formally, we focus on making predictions for any selected car and its interacting cars in next $T_f$ seconds $\hat{\textbf{Y}}_{t+1,t+T_f}$, based on observations of last $T_h$ seconds $\textbf{O}_{t-T_h,t}$:
\begin{equation}
    \hat{\textbf{Y}}_{t+1,t+T_f} = f(\textbf{O}_{t-T_h,t}).
\end{equation}
Specifically, we first make intention prediction with a high-level semantic graph network (SGN), which takes in the semantic graph (SG) $\mathcal{G}_{t - T_h, t}$ extracted from raw observation $\textbf{O}_{t-T_h,t}$, and predicts the goal state $g_t$:
\begin{equation}
    \mathcal{G}_{t - T_h, t} = f_{SG}(\textbf{O}_{t - T_h, t}).
\end{equation}
\begin{equation}
    g_t = f_{SGN}(\mathcal{G}_{t - T_h, t}).
\end{equation}
The goal state $g_t$ is then delivered as the intention signal along with historic dynamics $\textbf{S}_{t - T_h, t}$ to the low-level encoder-decoder network (EDN) for trajectory prediction:
\begin{equation}
    \hat{\textbf{Y}}_{t + 1, t + T_f} = f_{EDN}(\textbf{S}_{t - T_h, t}, g_t, \theta).
\end{equation}
To endow the trajectory prediction model with the capability to capture individual and scenario differences, we set up an online adaptation module, where a modified Extended Kalman Filter (MEKF$_\lambda$) algorithm is utilized to adapt the parameter of the model in real time. Specifically, we regard EDN as a dynamic system and estimate its parameter $\theta$ by minimizing the error between ground-truth trajectory in past $\tau$ steps $\textbf{Y}_{t-\tau}$ and predicted trajectory $\tau$ steps earlier $\hat{\textbf{Y}}_{t-\tau,t}$:
\begin{equation}
    \theta_{t} = f_{MEKF_{\lambda}}(\theta_{t-1}, \textbf{Y}_{t-\tau, t}, \hat{\textbf{Y}}_{t-\tau,t}).
\end{equation}

\section{Methodology}
Our model consists of a high-level intention prediction model, a low-level trajectory prediction model, and an online adaptation module. 
\subsection{Intention Prediction}
During driving in intense-traffic environment, humans will intentionally identify which slot is spatially and temporally suitable to insert into. Then a determined slot and associated goal position is generated, which is then used to guide low-level actions such as acceleration or steering. Based on this insight, we first adopt the dynamic insertion areas (DIA) to describe the slots on the road, which are regarded as nodes to construct a semantic graph (SG) $\mathcal{G}_{t - T_f, t}$. With SG, a semantic graph network (SGN) is then adopted to conduct relationship reasoning among the nodes in the graph and generate future goal state $g_t$. Detailed introduction on semantic graph and semantic graph network can be found at \cite{hu2020scenario, wang2021hierarchical}.

\begin{algorithm}[t]
	\caption{$\tau$-step online adaptation with MEKF$_{\lambda}$}
	\label{alg:adaptation}
		\textbf{Input:}
		Offline trained EDN network with parameter $\theta$, initial variance $\textbf{Q}_t$ and $\textbf{R}_t$ for measurement noise and process noise respectively, forgetting factor $\lambda$\newline
		\textbf{Output:}
		A sequence of generated future behavior $\{\hat{\textbf{Y}}_{t}\}_{t=1}^T$
	\begin{algorithmic}[1]
		\For{$t=1, 2, ..., T$} 
		    \If {$t\geq\tau$}
		        \State stack recent $\tau$-step observations:
		       \State\hspace{2em}$\textbf{Y}_{t-\tau, t} = [y_{t-\tau+1}, ..., y_{t}]$
		        \State stack recent $\tau$-step generated trajectory:
		         \State\hspace{2em}$\hat{\textbf{Y}}_{t-\tau, t} = [\hat{y}_{t-\tau+1}, ..., \hat{y}_{t}]$
		        \State adapt model parameter via MEKF$_{\lambda}$: 
		        
		        \State\hspace{2em}$\textbf{H}_t = \frac{\partial\hat{\textbf{Y}}_{t-\tau,t}}{\partial \hat{\theta}_{t-1}}$
		        
		      \State\hspace{2em}$\textbf{K}_t = \textbf{P}_{t-1}\cdot\textbf{H}_t^T\cdot(\textbf{H}_t\cdot\textbf{P}_{t-1}\cdot\textbf{H}_t^T+\textbf{R}_t)^{-1}$
		      
		      \State\hspace{2em}$\textbf{P}_t=\lambda^{-1}(\textbf{P}_t-\textbf{K}_t\cdot\textbf{H}_t\cdot\textbf{P}_{t-1}+\textbf{Q}_t)$
		      
		      \State\hspace{2em}$\hat{\theta}_{t} = \hat{\theta}_{t-1} + \textbf{K}_t \cdot (\textbf{Y}_{t-\tau,t} - \hat{\textbf{Y}}_{t-\tau,t})$
		    \Else
		        \State initialization: $\hat{\theta}_t=\theta$
		    \EndIf
		    \State collect goal state $g_t$ from SGN and historic states $\textbf{S}_t$.
		    \State generate future behavior:
		        \State\hspace{2em}$\hat{\textbf{Y}}_{t}=[\hat{y}_{t+1}, ...\hat{y}_{t+T_f}]=f_{EDN}(\textbf{S}_t, g_t, \hat{\theta}_t)$.
    	 \EndFor
	 \State\Return sequence of predicted behaviors $\{\hat{\textbf{Y}}_{t}\}_{t=1}^T$
	\end{algorithmic}
\end{algorithm}

\subsection{Trajectory Prediction}
To generate future behaviors of arbitrary length with sufficient expressiveness, we use the encoder-decoder network (EDN) \cite{cho2014learning,neubig2017neural} as the low-level trajectory prediction policy. The EDN consists of two GRU networks, namely the encoder and the decoder. At any time step t, the encoder takes in the sequence of historic states $\textbf{S}_t=[s_{t-T_h},..., s_t]$ and compresses all these information into a context vector $c_t$:
\begin{equation}
\label{eq:encoder}
	c_t = f_{enc}(\textbf{S}_t; \theta^{E}),
\end{equation}
where $\theta^{E}$ denotes the parameter of encoder. The context vector $c_t$, the current state $s_t$, and the goal state $g_t$ generated by high-level policy are then fed into the decoder to recursively generate future behavior predictions: 
\begin{equation}
	\hat{\textbf{Y}}_t =[\hat{\textbf{y}}_{t+1}, ..., \hat{\textbf{y}}_{t+T_f}] = f_{dec}(\textbf{c}_t, s_t, g_t; \theta^{D}),
\end{equation}
where $\theta^{E}$ denotes the parameter of encoder. Specifically, the decoder takes vehicle’s current state $s_t$ as the initial input to generate the first-step trajectory. In every following step, the decoder takes the output value of last step as the input to generate a new output. In this paper, we choose graph recurrent unit (GRU) as the basic RNN cell and stack three dense layers on the decoder for stronger decoding capability. According to the empirical results in \cite{wang2021hierarchical}, we append the goal state feature $g_t$ to the end of original input feature vector. The loss function is designed to can be simply designed minimize the error between the ground-truth trajectory and the generated trajectory:

\begin{equation}
    \mathcal{L} = \sum_{i=0}^N||\hat{\textbf{Y}}_i - \textbf{Y}_i||_2,
\end{equation}
where $N$ denotes the number of training trajectory samples.

\begin{figure}[t!]
    \centering
    \includegraphics[width=0.4\textwidth]{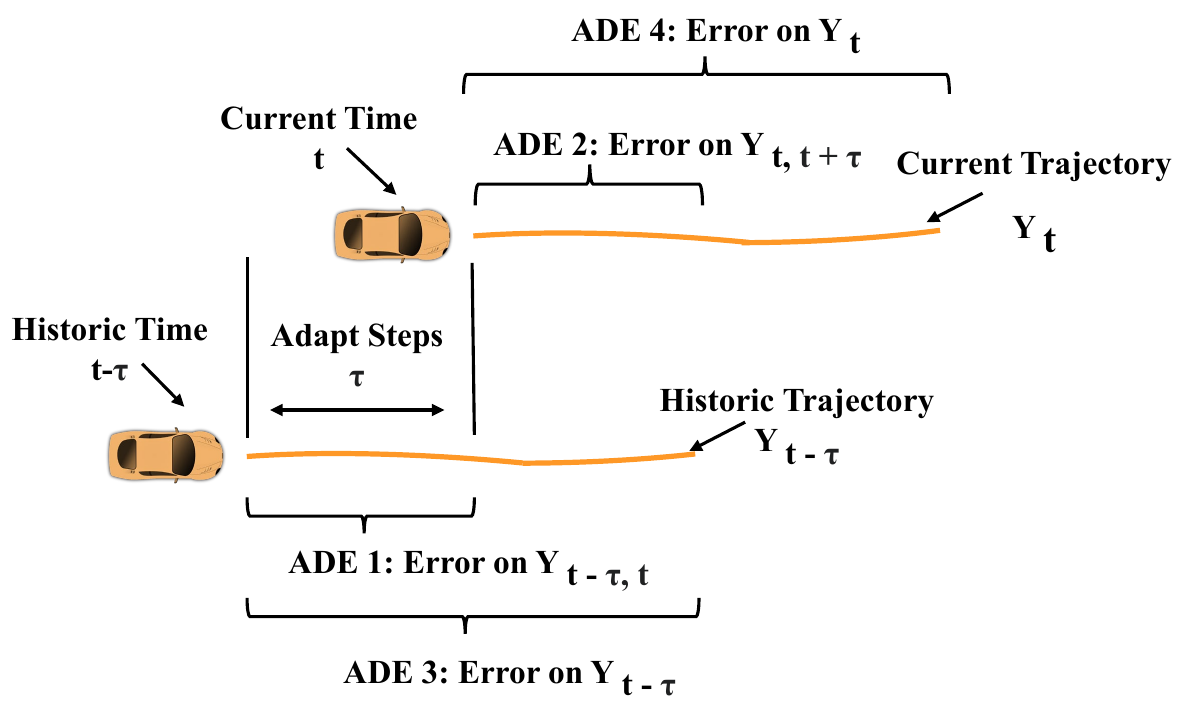}
    \caption{An illustration for online adaptation and proposed 4 metrics for performance evaluation. At time step $t$, the model parameters can be adapted by minimizing prediction error of the trajectory in past $\tau$ steps $\textbf{Y}_{t-\tau, t}$. The adapted parameter is then used to generate prediction $\textbf{Y}_t$ on current time $t$. 4 metrics are designed to evaluate the adaptation performance. ADE 1 can verify how the adaptation works on the source trajectory $\textbf{Y}_{t-\tau, t}$. ADE 2 can verify whether adaptation can benefit short-term prediction in the presence of behavior gap between earlier and current time. ADE 3 verifies whether we have obtained enough information from the source trajectory $\textbf{Y}_{t-\tau, t}$. ADE 4 verifies whether adaptation can benefit long-term prediction at current time $t$.}
    \label{fig:adaptation_trajectory}
\end{figure}

\subsection{Online Adaptation}
Adaptable prediction to individuals requires the adaptation of parameters in the neural network in real time. Due to the huge time delay in obtaining ground-truth intention label, in this paper, we only consider the online adaptation for the low-level behavior-prediction policy (EDN). Formally, we regard the adaptation of the neural network as a parameter estimation process of a nonlinear system with noise:


\begin{figure}[t!]
    \includegraphics[width=0.45\textwidth]{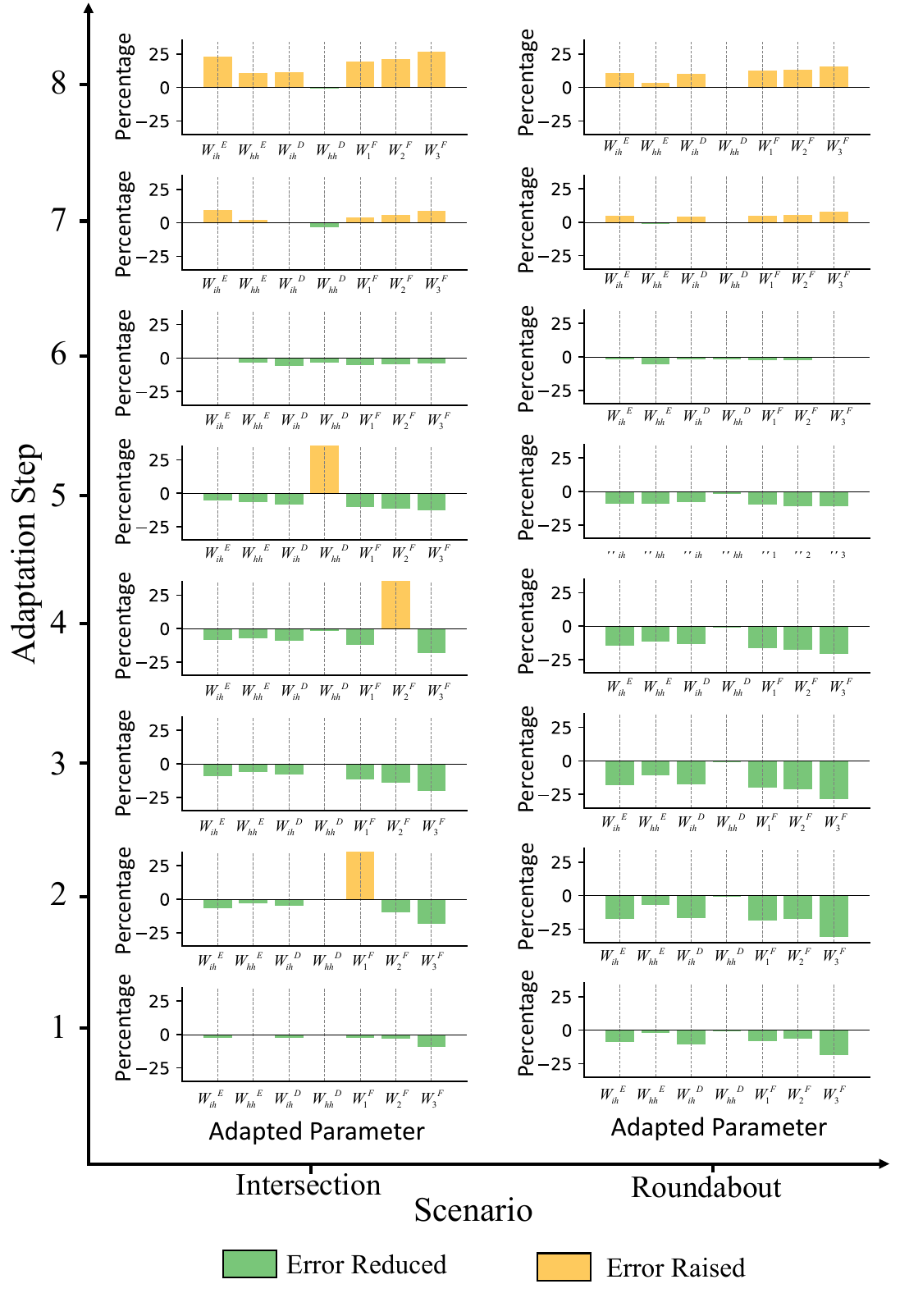}
    \caption{Percentage of error (ADE 2) reduced or raised after online adaptation, under 1) different adaptation step $\tau$; 2) different layer of parameters adapted; 2) different scenarios. Three conclusions from the results: 1) online adaptation works the best around the adaptation step of 2 or 3; 2) prediction accuracy was improved by a higher percentage in the roundabout scenario (transferred) than in the intersection scenario (trained); 3) the best adaptation performance is obtained usually by adapting $W^F_3$, the last layer of the FC network in the decoder.}
    \label{fig:adaptation_layers}
\end{figure}

\begin{figure*}[t!]
    \centering
    \includegraphics[width=0.71\textwidth]{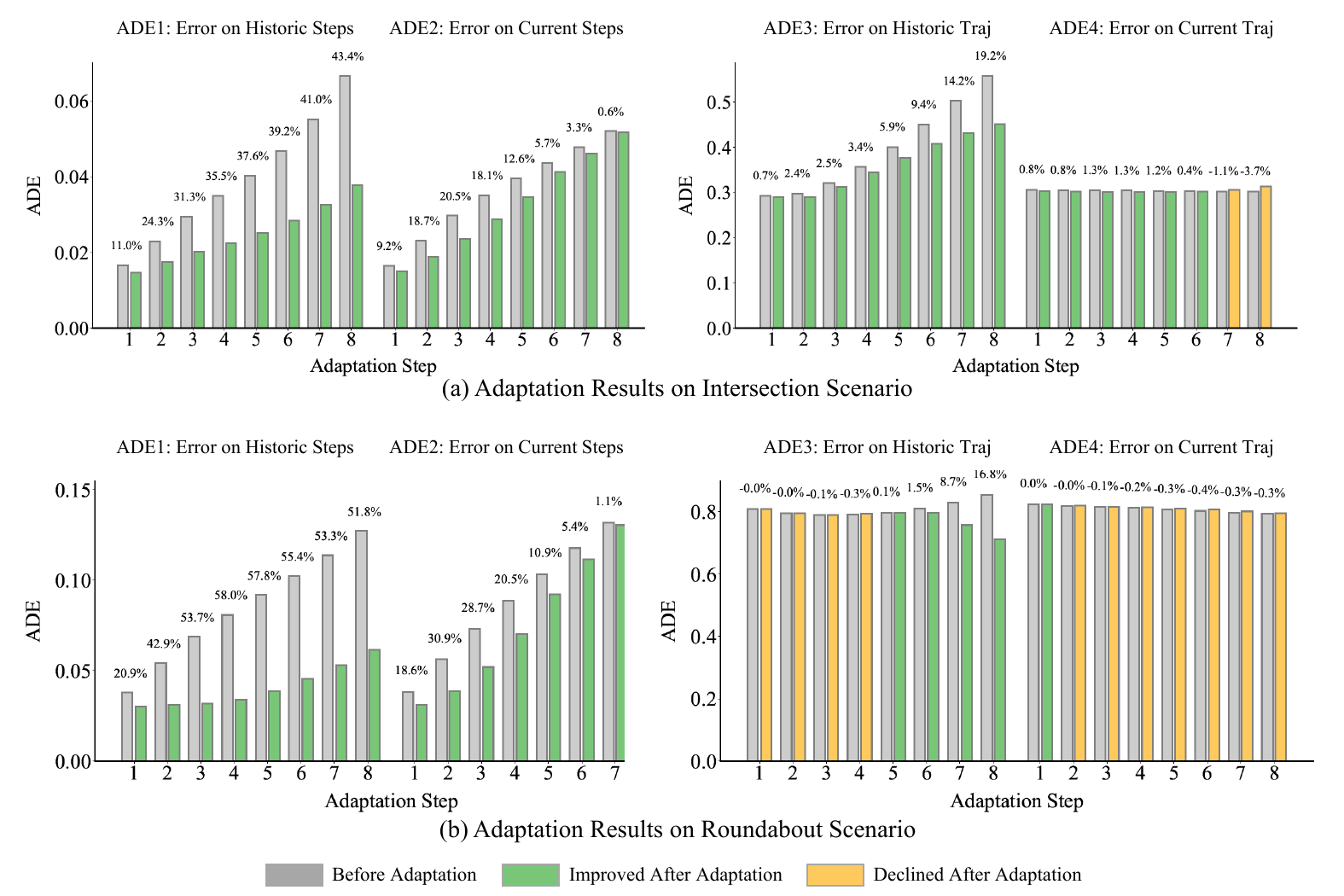}
    \caption{Online adaptation performance analysis. 1) ADE 1 and ADE 3: as adaptation step $\tau$ increased, the online adaptation could get more information and improve prediction accuracy by a higher percentage. 2) ADE 2: as $\tau$ increases, the improved percentage first grew higher due to more information obtained, but then declined due to the behavior gap between earlier time and current time. When $\tau$ was 3, the short-term prediction was improved by 20\% on intersection scenario (customizability) and 28\% on roundabout scenario(transferability). 3) ADE 4: online adaptation can barely benefit long-term prediction.}
    \label{fig:adaptation_analysis}
\end{figure*}

\begin{equation}
    \textbf{Y}_{t} = f_{EDN}( \textbf{S}_t, g_t, \hat{\theta}_{t}) + \textbf{u}_t    
\end{equation}
\begin{equation}
    \hat{\theta}_{t} = \hat{\theta}_{t-1} + \omega_{t}
\end{equation}
where $\theta$ denotes the parameter of EDN; the measurement noise $u_{t}\sim \mathcal{N}(0, \textbf{R}_t)$ and the process noise $\omega_{t}\sim \mathcal{N}(0, \textbf{Q}_t)$ are assumed to be Gaussian with zero mean and white noise. For simplicity, we assume $\textbf{Q}_t=\sigma_q\textbf{I}$ and $\textbf{R}_t=\sigma_r\textbf{I}$ where $\sigma_q>0$ and $\sigma_r>0$.

A modified Extended Kalman Filter (MEKF$_\lambda$) algorithm is then developed as in Algorithm~\ref{alg:adaptation} to make linearization of the model and adapt arbitrary layer in the neural network. To realize feedback decay from older observations, we modified EKF to include the forgetting factor $\lambda$. Note that a multi-step adaptation strategy is considered, meaning that prediction error of $\tau$ step is taken as the feedback to adapt the model.

\subsection{A new set of metrics}
As in Figure \ref{fig:adaptation_trajectory}, we proposed a new set of metrics to systematically analyze the performance of online adaptation:
\begin{enumerate}
    \item ADE 1: This metric evaluates the prediction error of the adapted steps on the historic trajectory $Y_{t-\tau, t}$. Because these steps are the observation source used to conduct online adaptation, the metric can verify whether the algorithm is working or not.
    \item ADE 2: This metric evaluates the prediction error of the adaptation steps on the current trajectory, which aims at verifying how the time lag is influencing the prediction. Also, this method can be used to verify whether adaptation could improve short-term behavior prediction.
    \item ADE 3: This metric evaluates the prediction error of the whole historic trajectory, which shows if we have gotten enough information on the behavior pattern.
    \item ADE 4: This metric evaluates the prediction performance of the whole current trajectory, which shows whether or not the adaptation based on historic information can help current long-term behavior generation.
\end{enumerate}

\section{Experiment}
\subsection{Experiment setting}
We verified our method with real driving data from INTERACTION dataset~\cite{interactiondataset}. Our model is trained on the MA intersection scenario by splitting 19084 data points to 80\% of training data and 20\% of testing data. The model is then directly tested on the FT roundabout scenario with 9711 data points to evaluate the transferability. In our experiments, we choose the historic time steps $T_h$ as 10 and future time step $T_f$ as 30 with frequency of 10Hz. We utilized Adam optimizer and sweep over more then 20 combinations of hyperparameter for best performance. We proposed a new set of 4 metrics to systematically evaluate the performance of online adaptation, as show in Figure~\ref{fig:adaptation_analysis}.

\subsection{The best layer to adapt}
There exist many layers in the END. We denote the encoder GRU’s input-hidden weights as $W^E_{ih}$, the encoder GRU’s hidden-hidden weights as $W^E_{hh}$, the decoder GRU’s input-hidden weights as $W^D_{ih}$, the decoder GRU’s hidden-hidden weights as $W^D_{hh}$, and the weights of the three-layer FC stacked on top of decoder as $W^F_1$, $W^F_2$, $W^F_3$ respectively. While it remains theoretically unknown which layer is the best to adapt due to the lack of interpretability of neural networks, as in Figure~\ref{fig:adaptation_layers} we empirically found that $W_3^F$, the last layer the FC network in the decoder, is the best to adapt.

\subsection{Trade-off on adaptation step $\tau$}
When we increase the adaptation step $\tau$, there is a trade-off between how much information we can get and the behavior gap between earlier time $t-\tau$ and current time $t$. Figure~\ref{fig:adaptation_analysis} shows our exploration on the best step $\tau$ to adapt. Several observation can be drawn: 1) prediction performance decayed when the model was directly transferred to the roundabout scenario. 2) online adaptation could benefit short-term prediction as in ADE 2, and could barely improve long-term predcition as in ADE 4. 3) When the $\tau$ increases the improved percentage in ADE 2 first rose, reached peak $\tau$ of 2 or 3, and then dropped. Specifically, the ADE 2 is reduced by 20\% in the intersection scenario (customizability), and by 28 \% in the roundabout scenario (transferability).

\section{Conclusion}
This paper extended a hierarchical prediction model to include adaptability, where MEKF$_\lambda$ was used to adapt to different human drivers and driving scenarios. A new set of metrics is proposed to systematically evaluate the performance of online adaption. Our experiments demonstrated online adaptation could improve short-term prediction accuracy by 20\% and 28\% in the trained and transferred scenario respectively. We also empirically found the best choice of layer is the last layer of FC in the decoder, and the best adaptation step $\tau$ is around 2 or 3. Future efforts include online adaptation of intention predictions, online adaptation on more driving scenarios, and interpretation of neural network.




\bibliography{aaai22}

\appendix

\end{document}